# Strategy for Rapid Diabetic Retinopathy Exposure Based on Enhanced Feature Extraction Processing

V. Banupriya[1,*] and S. Anusuya[2]

[1]Department of Computer Science and Business Systems, M. Kumarasamy College of Engineering, Karur, 639113, Tamilnadu, India
[2]Department of Information Technology, Saveetha School of Engineering, Saveetha Institute of Medical and Technical Sciences, SIMATS, Chennai, 602117, Tamil Nadu, India
*Corresponding Author: V. Banupriya. Email: banucs03@gmail.com

**Abstract:** In the modern world, one of the most severe eye infections brought on by diabetes is known as diabetic retinopathy (DR), which will result in retinal damage, and, thus, lead to blindness. Diabetic retinopathy (DR) can be well treated with early diagnosis. Retinal fundus images of humans are used to screen for lesions in the retina. However, detecting DR in the early stages is challenging due to the minimal symptoms. Furthermore, the occurrence of diseases linked to vascular anomalies brought on by DR aids in diagnosing the condition. Nevertheless, the resources required for manually identifying the lesions are high. Similarly, training for Convolutional Neural Networks (CNN) is more time-consuming. This proposed research aims to improve diabetic retinopathy diagnosis by developing an enhanced deep learning model (EDLM) for timely DR identification that is potentially more accurate than existing CNN-based models. The proposed model will detect various lesions from retinal images in the early stages. First, characteristics are retrieved from the retinal fundus picture and put into the EDLM for classification. For dimensionality reduction, EDLM is used. Additionally, the classification and feature extraction processes are optimized using the stochastic gradient descent (SGD) optimizer. The EDLM's effectiveness is assessed on the KAGGLE dataset with 3459 retinal images, and results are compared over VGG16, VGG19, RESNET18, RESNET34, and RESNET50. Experimental results show that the EDLM achieves higher average sensitivity by 8.28% for VGG16, by 7.03% for VGG19, by 5.58% for ResNet18, by 4.26% for ResNet 34, and by 2.04% for ResNet 50, respectively.

## 1 Introduction

Diabetes-related eye illness is known as diabetic retinopathy (DR), contributes to retinopathy. The DR seems forbidding and will progress over time. It will damage the eye's retina and gradually result in vision impairment. Individuals with long-term, weakly-delimited diabetes mellitus are more likely to be affected by DR. The DR-impacted patients will often suffer from symptoms like blurred and impaired color vision, fluctuating vision, and spots in the vision. Diabetes will impulse damage to the retina's blood vessels, and blood leakage inside the vessel will result in the occurrence of non-proliferative DR [1]. The DR stage with severity will have abnormal growth of new blood vessels, micro-aneurysms, exudates, and hemorrhages. This will occur as the blood is unable to spurt inside the retina. In the proliferative DR stage, new growth of delicate blood vessels is observed due to the lack of oxygen in the retina. These vessels may bleed, resulting in retinal degeneration and vision loss. Fig. 1 will depict the changes in the retina due to DR.

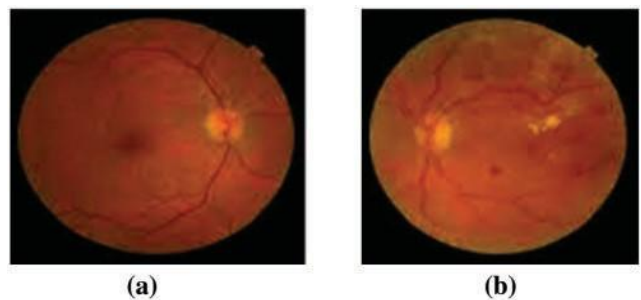

**Figure 1:** (a) Normal retinal image and (b) Diabetic retinopathy image

Being a progressive disease, DR has various levels of severity that are categorized as follows: No DR, mild Nonproliferative DR (NPDR), moderate NPDR, severe NPDR, and proliferative DR. Based on retinal clinical findings, DR is categorized in Table 1 for comparison. The DR's severity level for each stage primarily depends on specific features such as hemorrhages, micro-aneurysms, exudates, and "cotton wool" spots, which can be observed in the retina. These features will be absent in a no-DR-stage retina. The non-proliferative DR category will constitute the mild, moderate, and severe stages. In a mild DR stage, the micro-aneurysms will expand, covering the limited space inside the blood vessel. The moderate DR stage will have a stoppage of blood flow through the vessels and a loss in the ability to transfer blood to the retina. Due to this, there will be the appearance of numerous micro-aneurysms along with hemorrhages [2].

Table 2 represents the NPDR grading depending on the amount of Microaneurysms (MA) in the retinal image. For Normal MA count is 0, mild has an MA count between 0 to 5, Moderate has an MA count between 5 and 15 without neovascularisation, and an MA count exceeding 15 may represent the severe grading.

Fig. 2 represents the retinal fundus images with DR-related lesions like hemorrhages, micro-aneurysms, and hard and soft exudates. These images are employed for diagnosing an extensive range of eye diseases. The images include DR, glaucoma, cataracts, and retinopathy of prematurity. Identifying diseases using image processing and machine learning methods is effective [3]. For the identification of DR, fundus images were grabbed by a 2D monocular camera. Contrary to this scanning of another eye, which seems like the Optical Coherence Tomography (OCT) images, and the angiographs, the acquisition of fundus images is relatively cost-effective as well as non-invasive. Hence, it exhibits feasibility enough for heavy-range testing [4]. Deep learning [5], a subset of machine learning methods, will involve stratified layers of processing phases with unsupervised characteristics learning and classification pattern analysis. Deep learning is one of the computer-aided medical diagnosis methods. Of late, there has been extensive utilization of deep learning for DR detection as well as classification. It can grasp the provided data's characteristics despite the integration of multiple heterogeneous sources.

**Table 1:** Diabetic retinopathy classification based on clinical retinal findings

| DR level | Diagnosis | Clinical findings |
|---|---|---|
| Level 0 | With no DR | Abnormalities-NIL |
| Level 1 | With mild NPDR | Only microaneurysms |
| Level 2 | With moderate NPDR | Hard exudates (HE) Hemorrhages (H), Microaneurysms (MA), |
| Level 3 | Severe NPDR | Each of the four quadrants should have a value of 20 H, venous bleeding in two quadrants. |
| Level 4 | Proliferative DR | Vitreous haemorrhage |

**Table 2:** The NPDR grades

| Diabetic retinopathy grading | Microaneurysms count |
|---|---|
| Normal DR | 0 |
| Mild DR | >0 and <5 |
| Moderate DR | >5 and < 15 (without neovascularisation) |
| Severe DR | >15 |

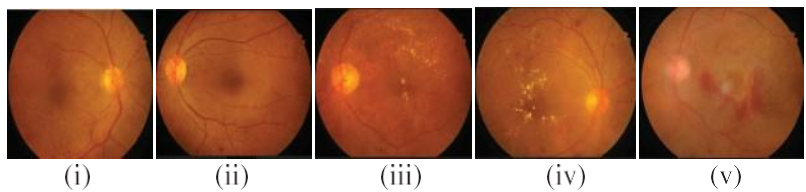

(i)　　(ii)　　(iii)　　(iv)　　(v)

**Figure 2:** Stages of DR (i) No DR, (ii) mild NPDR, (iii) moderate NPDR, (iv) severe NPDR, (v) proliferative DR

CNNs comprise various image applications, analyses, and interpretations, including medical imaging. In the 1970s, there was the routine construction of network architectures that were designed to work with image data and practical applications and also outperformed other methods concerning complex, challenging tasks such as recognizing handwritten characters. Nevertheless, the feasibility of the CNNs increased for more complicated problems of image recognition problems only after many advancements in neural networks like the deployment of dropout associated with the increase of exponential power employing the Graphical Processor Units (GPUs). Nowadays, there is the utilization of enormous CNNs for successful management, with significant complexity in recognizing the phases of multiple object classes [6]. Automated grading, or, to be more specific, CNNs, suffer from two key issues [7]. The first issue is associated with the accomplishment of a desirable offset in the sensitivity as well as the specificity (i.e., there is a proper determination of the patients with non-DR). This is substantially tougher for a five-class problem. Yet another issue faced by neural networks is over-fitting. The skewed datasets will result in the network overfitting to the dataset's most prominent class [8]. Usually, the vast datasets are tremendously skewed where a couple of classes may have less than 3% of images, which implies that the network had to have changes to ensure its capability to continue learning these images' features.

Main contributions of our work:

- The proposed work involves early predicting of diabetic retinopathy (DR) among young humans, including both genders.
- The subject selection of the current work involves not only the prediction of diabetic retinopathy but also identifying the lesions. Predictions such as MA, and Haemorrhages in the retinal fundus imagery for classifying Diabetic Retinopathy (DR) based on their severity.
- The selected Retinal fundus images of the subjects have undergone a Hybrid filter with Enhanced Contrast Limited Adaptive Histogram Equalization systematically to obtain optimum values for enhancing the quality of retinal images without losing their fine details.
- The current work involves designing a fully Automated Enhanced Deep Neural Network Model (EDL Model) to categorize and recognize retinal images.
- The EDLM in this current work is tuned with discriminative features using kernel filtering.

This research article follows a chronicle structure. The Literature survey based on the Early prediction of DR is discussed in Section 2; The methodology carried out in developing the Diabetic Retinopathy classification Model is explained in Section 3; Section 4 discusses the study's findings using quality assessment metrics; The Conclusion of the entire work is given in Section 5.

## 2 Related Work

Identifying diseases using image processing and machine learning methods is effective [9]. For the identification of DR, fundus images were grabbed by a 2D monocular camera. Contrary to this scanning of another eye, which seems like the OCT images, and the angiographs, the acquisition of fundus images is relatively cost-effective as well as non-invasive. A paradigm that offers a protective approach against the adversarial backscatter attack, the antagonistic training, and a matching feature technique that maintains the categorization with accurate labeling is introduced by Lal et al. in [10]. For the Diabetic Retinopathy detection task, which would be regarded as a cutting-edge attempt, we examine and analyze the antagonistic attacks as well as countermeasures on the retinal fundus images. Observations on antagonistic attack-prone retinal fundus images show that the suggested protective model is trustworthy with 99% accuracy.

A novel technique for precise bleeding diagnosis via retinal fundus pictures is put forth by Maqsood et al. [11]. The boundary characteristics from the source retinal fundus images are first improved by the suggested method using the enhanced visibility restoration method. Then, a convolutional neural network (CNN) architecture is indicated in the second level to find hemorrhages. Finally, the identified hemorrhages are used to obtain features from a customized pre-trained CNN model. The convolutional dense visual wavelet transform combines all retrieved relevant features inside the third stage. Then the features are chosen using a multi-logistic extrapolation regulated entropy change of scenery. Paradise et al. [12] employed a CNN architecture known as the ResNet-50 to extract features for further classification. The ResNet-50 feature output was also employed as the input for machine learning (ML) classifiers like Support Vector Machine (SVM), and Random Forest (RF).

The proposed model used the DIARETDBI dataset's fundus images. This work puts forward data augmentation as well as pre-processing to enable the model for image recognition. Sensitivity and accuracy metrics were utilized to assess each one of the classifier's performance. For example, the maximum precision of 100% was achieved by the K-Nearest Neighbor (k-NN) classifier, while the precision of the SVM classifier was 99%.

Shanthini et al. [13] introduced a baseline detection technique for DR segmentation that included the classification of the retinal image's input feature with its foreground as well as backend. They then would handle them via segmentation at the pixel level Segmentation. The procedure of layer assessment was supplemented with the utilization of a two-layer CNN. The analysis of the false positives at the time of classifying the features. The proposed model sequentially produces results to identify the distinct infected region of the retina. The proposed method was relatively stable as it was able to achieve better accuracy of detection, sensitivity, as well as true positives. Mule et al. [14] utilized the DR dataset from Kaggle. The input image set was classified with the deep CNNs. For the experimentation, there was a preference for the Dense Net, the ResNet 50, and the VGG-16 architectures. The deployment of image processing methods such as resizing, the Gaussian blur technique, cropping, and so on. It was evident from the experimental results that the VGG-16 had better performance with an accuracy of 77.49% in comparison to the DenseNet as well as the ResNet 50 for the DR detection.

Martinez-Murcia et al. [15] presented a computer-aided diagnosis tool with deep learning methods. The proposal was based on a deep residual CNN to extract discriminatory features without any prior complex image transformations. Furthermore, the transfer learning concept for reusing layers from the Deep Neural Networks (DNN) that were earlier learned with the ImageNet dataset suggests. Under certain hypothetical conditions, the top layers, which had captured the abstract features, could be rebuilt for diverse issues. Experiments were done with the distinct architecture of convolution techniques. Gayathri et al. [16] proposed an automated DR grading method. Also, they would perform categorization based on the severity with the utilization of deep learning as well as machine learning (ML) algorithms. The method utilized a multipath CNN (M-CNN) to extract local and global features from the fundus images. Following that, use a ML classifier to classify the input data based on its severity. The process was validated across various databases and several machine learning classifiers (e.g., SVM, RF, and J48). The experiments showed that the M-CNN network yielded the best response.

Automated grading, or, to be more specific, CNNs, suffer from issues such as Scarcity in labelled data, Class imbalance, Time-consuming to train CNN and an Optimal architecture framework. The first issue is associated with the accomplishment of a desirable offset in the sensitivity as well as the specificity (i.e., there is a proper determination of the patients with non-DR). This is substantially tougher for a five-class problem. Yet another issue faced by neural networks is over-fitting. The skewed datasets will result in the network overfitting to the dataset's most prominent class [10]. Usually, the vast datasets are tremendously skewed where a couple of classes may have less than 3% of images, which implies that the network had to have changes to ensure its capability to continue learning these images' features.

The CNN architecture depends on the problem to be solved. Pre-trained Models like VGG16, Inception, and ResNets can overcome sampling deficiencies as it is already trained with varying types of images and can identify a range of features. These Models usually have complex architectures as it is trained to differentiate thousands of classes. The benefit of the pre-trained classifier for making predictions about various entities can be an obstacle when it is required for specific purposes like DR identification, as the pre-trained deep learners can overfit the data. Thus, it is required to propose an enhanced Deep Learning Model which is suitable for the early Diabetic Retinopathy (DR) identification.

## 3 Proposed Methodology

The proposed diabetic retinopathy classification uses retinal fundus images downloaded from the KAGGLE and IRDiD datasets. Initial pre-processing of the image involves using hybrid filtering and amplified Contrast-limited adaptive histogram equalization (CLAHE) with a range of CLAHE limitations. A strategy for dynamic CNN training has been presented in which there is a dynamic selection of the informative standard samples at each training epoch from massive medical images. Each pixel throughout the false negative stream has dynamic weights assigned to it to show how informative it is. The significance of every negative test pixels should be updated following each CNN training session. Until a stopping criterion is met, this procedure will be repeated. A pixel likelihood map will be produced for each test image as a result of the training set being used to categorize each pixel in the testing set. This section will offer discussions on the CNN image segmentation, the VGGNet, and the ResNet DL Models.

The primary goal of our model is to increase the detection of DR in retinal fundus images. The outcomes are contrasted with various CNN models. Performance evaluation of the proposed EDLM is evaluated with various metrics. The Kaggle Retinal Fundus Pictures dataset consists of 54,376 testing pictures with an undetermined DR level. The 33,136 images for training have been categorized into five stages of diabetic retinopathy. Acquisition of the images was made with many fundus cameras as well as diverse fields of view [17]. We selected a subset of 3549 retinal images from the Kaggle dataset.

### Pre-Processing

For improved prediction accuracy, the contrast of retinal pictures is increased using a hybrid image filter and enhanced CLAHE. In order to avoid processing a picture with artifacts caused by excessive edge detection, the CLAHE trimmed the histogram at a predetermined cutoff threshold [18]. Fig. 3 represents the proposed preprocessing model. It is possible to resolve this issue by applying the enhanced CLAHE. There will be the generation of a look-up table. These values will form the basis for generating a histogram for the input image. The clip limit value is employed for clipping the histogram, which will vary from 0.002–0.005. Afterward, there will be a mapping of the histogram. In the end, there will be image interpolation. The fixed clipping feature can be overcome with the enhanced CLAHE outcome. The outcome of this enhanced Preprocessing with hybrid filter and Enhanced CLAHE results improved retinal image quality.

### EDLM Framework

The Proposed Enhanced Deep Learning Model architecture comprises five distinct convolutional layers (CL) followed by the ReLUs and the spatial max-pooling as displayed in Fig. 4. The network's final layers will contain an final soft-max classification layer (FCL) [19–21]. In every convolution layer, 32 small-size filters are used. After the CL, the feature map sizes are split in half using a cadence of size $2 \ast 2$. Max-pooling (size $2 \ast 2$) will include small spatial invariance within the network, and fewer free parameters will be used. There is an addition of weight decay to every layer to penalize the vast weight parameters during the gradient's back-propagation in the optimization routine. Most of the designs get started from standard methods and get optimized. We used VGG 16 Network with 14 million parameters to be optimized, and based on the VGG16 network framework, we further optimized it by reducing the number of layers and hyperparameter tuning. The proposed model has 20 layers. It is

CMC, 2023, vol.75, no.3          5603

based on a hypothesis that 20 layers give a better outcome, and it has been proved by comparing it with VGG16.

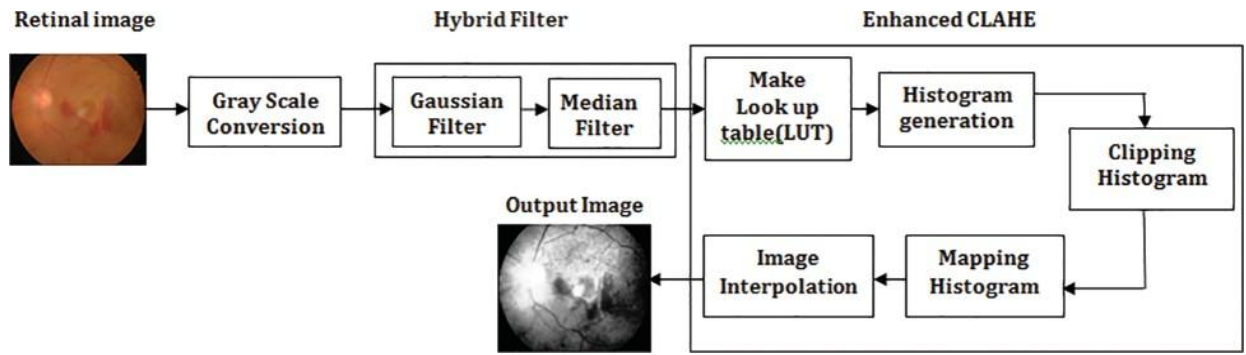

**Figure 3:** Proposed preprocessing model with hybrid filter and enhanced CLAHE

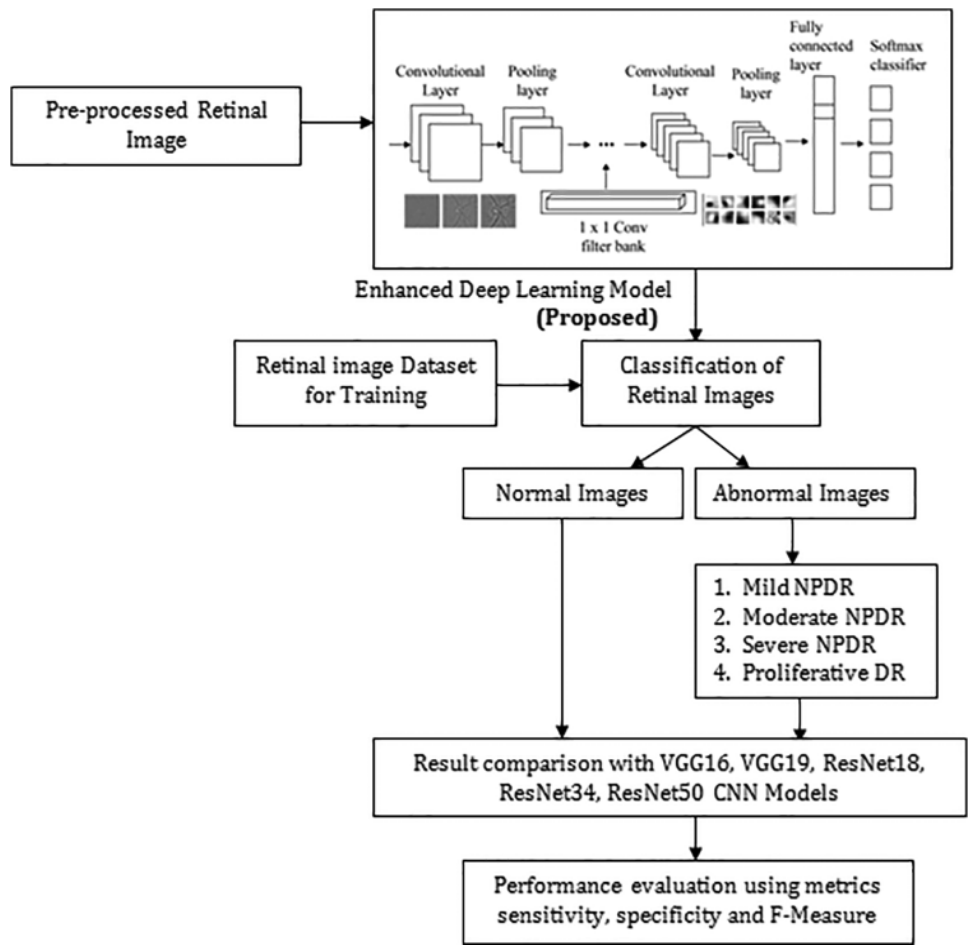

**Figure 4:** Proposed approach for predicting diabetic retinopathy

*Steps to Select the Hyper-Parameters*

Step 1. Split the data at hand into training and test subsets.

Step 2. Repeat the optimization loop a fixed number of times or until a condition is met:

Select a new set of model hyperparameters.

Step 3. Compare all metric values and choose the hyperparameter set that yields the best metric value. The typical standard hyperparameter for the learning rate is 0.001. The hyperparameter tuning method used is river formation dynamics.

*Algorithm of Proposed EDL Model*

---

**Input :** Preprocessed color retinal fundus image of pixel size 192 × 192 is given as input.
**Output :** Classification of Retinal fundus Image based on severity.

Step 1: The preprocessed image is fed into the convolutional layer that performs image features extraction.

$$\sigma_k^{(l)} = \sum_{i=1}^{i=k^{l-1}} x^{(l)_i}.w^{(l)_i} + b^{(k)}$$

Step 2: The feature map generated is then given to ReLu which results Rectified Feature map.

$$f(x) = \max(0, x), s_k^{(l)} = f\left(\sigma_k^{(l)}\right)$$

Step 3: Each rectified feature map's dimensionality is decreased by the max pooling

Layer.

$$l_k^{(l)} = pool\left(s_k^{(l)}\right)$$

Step 4: The resulting image features vector from convolutional and pooling layers is fed into FCL and used to categorize the input data images into different classes.

Step 5: Soft-max logistic regression will output a score which will range between 0 and 1, thus, denoting the pixel's probability in belonging to the positive class.

$$C(l, s) = -\sum_{i=0}^{B} l_i \log(s_i) + (1 - l_i) \log(1 - s_i)$$

---

The outcome of the proposed EDL Model Framework detects the Diabetic Retinopathy and classification is done based on its severity with improved accuracy. The architecture of proposed enhanced deep learning model with omission of the ReLUs is described in the Table 3.

**Table 3:** Architecture of the proposed EDLM

| Name | Layer type | Feature map size | Filter Kernel size | Number of stride | Number of padding |
|---|---|---|---|---|---|
| | | Input image | | | |
| Conv1 | Conv | 224 × 224 × 64 | 3 × 3 × 64 | 1 × 1 | 1 × 1 |
| | Conv | 224 × 224 × 64 | 3 × 3 × 64 | | |
| | Maxpool | 112 × 112 × 64 | 2 × 2 | 2 × 2 | 0 × 0 |

(Continued)

**Table 3:** Continued

| Name | Layer type | Feature map size | Filter Kernel size | Number of stride | Number of padding |
|---|---|---|---|---|---|
| Conv2 | Conv | 112 × 112 × 128 | 3 × 3 × 128 | 1 × 1 | 1 × 1 |
| | Conv | 112 × 112 × 128 | 3 × 3 × 128 | | |
| | Maxpool | 56 × 56 × 128 | 2 × 2 | 2 × 2 | 0 × 0 |
| Conv3 | Conv | 56 × 56 × 256 | 3 × 3 × 256 | 1 × 1 | 1 × 1 |
| | Conv | 56 × 56 × 256 | 3 × 3 × 256 | | |
| | Conv | 56 × 56 × 256 | 3 × 3 × 256 | | |
| | Maxpool | 28 × 28 × 256 | 2 × 2 | 2 × 2 | 0 × 0 |
| Conv4 | Conv | 28 × 28 × 512 | 3 × 3 × 512 | 1 × 1 | 1 × 1 |
| | Conv | 28 × 28 × 512 | 3 × 3 × 512 | | |
| | Conv | 28 × 28 × 512 | 3 × 3 × 512 | | |
| | Maxpool | 14 × 14 × 512 | 2 × 2 | 2 × 2 | 0 × 0 |
| Conv5 | Conv | 14 × 14 × 512 | 3 × 3 × 512 | 1 × 1 | 1 × 1 |
| | Conv | 14 × 14 × 512 | 3 × 3 × 512 | | |
| | Conv | 14 × 14 × 512 | 3 × 3 × 512 | | |
| | Maxpool | 14 × 14 × 512 | 2 × 2 | 2 × 2 | 0 × 0 |
| Fully connected | | 4096 | 1 × 1 × 4096 | | |
| Fully connected | | 3 | 1 × 1 × 3 | | |
| Soft-max activation function | | | | | |

*Convolutional Layer*

The convolutional layer (CL) aims to perform image feature extraction as displayed in Fig. 5. The spatial relations among the pixels are found by convolution using the input data matrix. Filters of varying sizes, called kernels, are used in every convolution for detecting features like edge, blur, etc. The filter slides over the input image, and the feature map as a matrix is acquired [22–25].

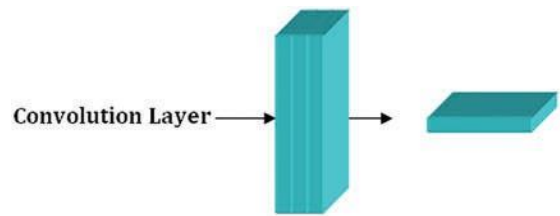

**Figure 5:** Convolution layer

In the first CL, the neuron weights serve as the filter values provided by:

$$\sigma_k^{(l)} = x_k^{(l)} * w_k^{(l)} + b_k^{(l)} \tag{1}$$

$$\sigma_k^{(l)} = \sum_{i=1}^{i=k^{l-1}} x_{k,i}^{(l)} * w_{k,i}^{(l)} + b_k^{(l)} \tag{2}$$

*ReLUs*

The Rectified Linear Units (ReLUs) function which is defined as $f(x) = max(0,\sigma)$ is used in this work. The outcome $s_k^{(l)}$ of a neuron $k$ and is defined as (3):

$$s_k^{(l)} = f\left(\sigma_k^{(l)}\right) \tag{3}$$

The ReLU function, when applied to the feature map, substitutes the negative pixel with zero. The resulting feature is called rectified feature map [26–28]. When the method is given a negative sign as input, it produces 0, but when it is given a positive sign, it yields x. The outcome, therefore, has a range of 0 to infinity.

*Max Pooling Layer and Feature Map*

A max pooling layer is a type of layer that typically follows a convolutional layer. The purpose of the max pooling layer is to reduce the spatial dimensions (width and height) of the feature map produced by the convolutional layer, while retaining the most important features. During max pooling, the input feature map is divided into non-overlapping rectangular regions. For each region, the maximum value is computed and the result is placed in the corresponding position of the output feature map. The size and stride of the pooling regions are typically chosen to downsample the feature map while preserving the spatial relationships between features. After maxpooling, a feature map is distinct in formulation (4).

$$l_k^{(l)} = pool\left(s_k^{(l)}\right) \tag{4}$$

*Fully Connected and Softmax Layer*

The resulting image featured a vector from convolutional, and pooling layers fed into FCL. The CNNs optimizes objective function, specifically Softmax function used. There is random initialization of all the network parameters in accordance with a normal distribution that has a variance which is equivalent to 0.05. The gradient descent, which has a parameter of $5.10^{-5}$ is used to train the CNN, and thus, will minimize a cost function which is expressed as per the below Eq. (5):

$$C(l,s) = -\sum_{i=0}^{B} l_i \log(s_i) + (1-l_i) \log(1-s_i) \tag{5}$$

Here, s denotes the probability score, l represents the reference label, and B indicates the samples [29]. The specification of a given epoch is 4000 mini-batches with a mini-batch length of 256 pixels. This means that approximately one million observations are used during CNN's learning, of which half are favorable and the other half is incorrect.

## 4 Stochastic Gradient Descent (SGD) Optimizer

A global optimization technique called gradient descent may find the optimal solutions to a number of problems. The basic idea is to constantly tweak the values to reduce the function f. A key Gradient Descent (GD) parameter is the stage count, which is determined by the scaling factor hyper-parameters [29–32]. It will take some time for the procedure to proceed via multiple repetitions until culmination. We run the danger of leaping the optimal value if indeed the repetition frequency is overly high. Stochastic refers to a procedure that has a variable possibility attached to it. As a consequence, only a few samples—rather than the full data set—are randomly selected for each cycle of SGD. The GD algorithm refers to "batch" as the collection of samples out of a database that are used to compute the gradient by each repeat. In typical GD methods like Batch GD, the batch is taken to be the complete dataset. Although using the complete dataset is tremendously beneficial for achieving the optimum in a lower noisy and much less arbitrary fashion, the problem arises when the sample size is huge. SGD processes one sample every cycle. To complete the iteration, the sampling is picked and blended at arbitrary.

---

**SGD Algorithm**

    Input data (samples, features)
      Set parameters' initial values to randomly selected values.
      weights = initialize (features)
      for i in range(iteration):
          Shuffle the data to introduce randomness data
          Loop through each sample in the shuffled data
      for j in range(samples):
          Select a single data point and its corresponding label
        Compute the predicted label for this data point
        Update the weights based on the difference between the predicted and actual labels
      Return the learned weights
    return weights

---

SGD uses the gradient of the minimization problem of a specific instance rather than the total of the gradients of all instances for each repetition. The strategy employed by the procedure to achieve the minima in SGD is typically noisier compared to that followed by a standard Gradient Descent algorithm since just one sampling from data is randomly selected for each iteration. But as long as we arrive at the minima with a noticeably lower training period, it doesn't really matter which route the method gives.

## 5 Results and Discussion

The proposed algorithm, the EDL Model Framework, has been tested over the quality of 3549 images, and several metrics have been assessed based on the test findings. The retinal scans obtained through Kaggle. Parameters for the experimental procedure is displayed in Table 4. Fig. 6 Shows an enhanced deep learning model to classify retinal pictures.

**Table 4:** Parameters taken into account for the experimental procedure

| Attributes | Features |
| --- | --- |
| Image dimensions | **192, 192, 3** |
| Total no. of retinal images tested | 3549 |
| 4–Proliferative DR | **252** |
| 3–Severe NPDR | **336** |
| 2–Moderate NPDR | **121** |
| 1–Mild NPDR | **672** |
| 0–Normal | 2168 |
| Image type | **Jpeg/png** |

| Input Image | Enhanced Preprocessing Model | Enhanced Deep Learning Model | Classification based on Lesions |
|---|---|---|---|
| 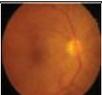 | 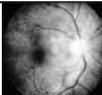 | *Normal Image* | *No abnormalities* |
| 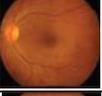 | 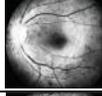 | *Abnormal Image* | *Microaneurysms (MA) only* |
| 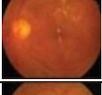 | 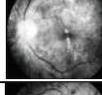 | *Abnormal Image* | *Haemorrhages* |
| 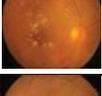 | 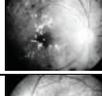 | *Abnormal Image* | *Hard Exudates* |
| 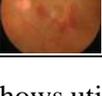 | 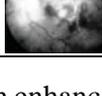 | *Abnormal Image* | *Neovascularisation* |

**Figure 6:** Shows utilizing an enhanced deep learning model to classify retinal picture**s**

The proposed Enhanced Deep Learning Model Framework is tested on various classes of retinal images normal images, mild NPDR, moderate NPDR, Severe NPDR and Proliferative DR images.

*Parameters for Performance Analysis*

In order to properly study picture categorization, the classifier's performance must be implemented. There are a number of metrics are employed to assess performance of classification. The proposed work measures metrics like Sensitivity, specificity, and F-Measure. The proposed EDML Framework is evaluated by,

True positive ($T_{posv}$),

False positive ($F_{posv}$),

True negative ($T_{negv}$) and

False negative ($F_{negv}$).

*Specificity*

The algorithm successfully assigns the suffix "specificity" to all healthy people. Clarity answers the following query: What percentage of healthy individuals did we anticipate accurately out of the total population?

$$\text{Specificity} = T_{negv} / T_{negv} + F_{posv} \tag{6}$$

*F-Measure*

The F1 Score considers both recall and precision. It is the accuracy and recall's harmonic mean. For the optimum F1 score, precision (p) and recall (r) should be balanced. On the other hand, if one metric is enhanced at the expense of another, the F1 score is lower. For instance, the F1 score is 0 if P is one and R is 0.

*Recall (Sensitivity)*

A recall is the proportion of those accurately identified as positive by our software among those with diabetes. Does recall offer an answer to the following query: How many diabetics did we correctly forecast out of all the persons with the disease?

$$\text{Recall (Sensitivity)} = T_{posv} / T_{posv} + F_{negv} \qquad (7)$$

- Numerator: +ve diabetic individuals.
- Denominator: All diabetes individuals

Fig. 7 illustrates the performance evaluation metrics of enhanced deep learning model framework on sensitivity, specificity, and F-measure. Table 5 compares the entire performance of the projected EDLM with various CNN models. The CNN models, like VGG16, VGG19, ResNet18, ResNet34, and ResNet50, are exploited for comparison. The data confirm that the suggested EDL model framework approach performed better than expected across all performance criteria, making it an effective tool for classifying DR.

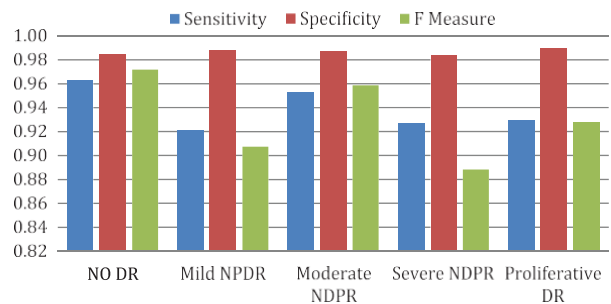

**Figure 7:** Performance evaluation metrics of EDLM framework

**Table 5:** Performance evaluation metrics of EDLM Framework on Sensitivity, Specificity, and F-Measure (Based on five classes of severity)

| Performance metrics | Image type | VGG16 | VGG19 | RESNET 18 | RESNET 34 | RESNET 50 | EDLM |
|---|---|---|---|---|---|---|---|
| Sensitivity | No DR | 0.91 | 0.92 | 0.93 | 0.94 | 0.95 | 0.96 |
|  | Mild NDPR | 0.83 | 0.85 | 0.87 | 0.88 | 0.90 | 0.92 |
|  | Moderate NDPR | 0.90 | 0.91 | 0.92 | 0.92 | 0.93 | 0.95 |
|  | Severe NDPR | 0.85 | 0.86 | 0.88 | 0.89 | 0.91 | 0.93 |
|  | Proliferative DR | 0.84 | 0.84 | 0.84 | 0.87 | 0.91 | 0.93 |
| Specificity | No DR | 0.97 | 0.97 | 0.97 | 0.98 | 0.98 | 0.98 |
|  | Mild NDPR | 0.97 | 0.97 | 0.98 | 0.98 | 0.98 | 0.99 |
|  | Moderate NDPR | 0.97 | 0.98 | 0.98 | 0.98 | 0.98 | 0.99 |
|  | Severe NDPR | 0.96 | 0.96 | 0.97 | 0.97 | 0.98 | 0.98 |
|  | Proliferative DR | 0.98 | 0.98 | 0.98 | 0.98 | 0.99 | 0.99 |
| F measure | No DR | 0.93 | 0.94 | 0.95 | 0.95 | 0.96 | 0.97 |
|  | Mild NPDR | 0.80 | 0.82 | 0.84 | 0.85 | 0.88 | 0.91 |
|  | Moderate NPDR | 0.92 | 0.92 | 0.93 | 0.93 | 0.94 | 0.96 |

(Continued)

**Table 5:** Continued

| Performance metrics | Image type | VGG16 | VGG19 | RESNET 18 | RESNET 34 | RESNET 50 | EDLM |
|---|---|---|---|---|---|---|---|
| | Severe NPDR | 0.74 | 0.77 | 0.80 | 0.81 | 0.85 | 0.89 |
| | Proliferative DR | 0.83 | 0.84 | 0.85 | 0.87 | 0.91 | 0.93 |

Fig. 8 illustrates the EDL model's increased average sensitivity by 8.28% for VGG16%, 7.03% for VGG19%, 5.58% for RESNET18%, 4.26% for RESNET34, and by 2.04% for RESNET50 respectively. Fig. 9 illustrates the EDL model's increased average specificity by 1.84% for VGG16%, 1.51% for VGG19%, 1.18% for RESNET18%, 0.94% for RESNET34, and by 0.51% for RESNET50 respectively.

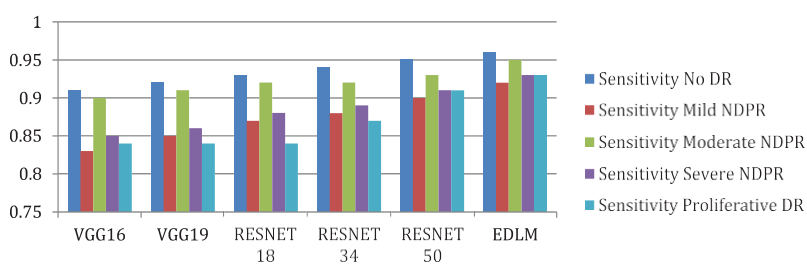

**Figure 8:** Comparison of sensitivity for enhanced deep learning model with other CNN models

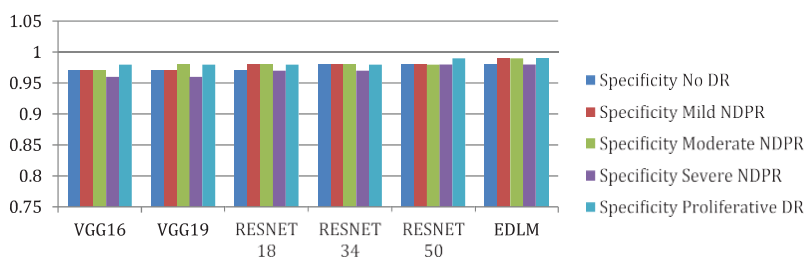

**Figure 9:** Comparison of specificity for enhanced deep learning model with other CNN models

Fig. 10 illustrates the EDL model's increased average f measure of 9.59% for VGG16%, 8.02% for VGG19%, 6.36% for RESNET18%, 5% for RESNET34%, and 2.62% for RESNET50. The proposed Enhanced Deep Learning Model architecture improves the classification of DR with high accuracy.

Table 6 compares the computational performance of the projected EDLM with various CNN models. The CNN models, like VGG16, VGG19, ResNet18, ResNet34, and ResNet50, are demoralized for assessment. The data confirm that the suggested EDL model framework approach performed better than expected across all CNN model. Overall, compared to other methods, the suggested one performs better at detecting the rapid vulnerability to diabetic retinopathy.

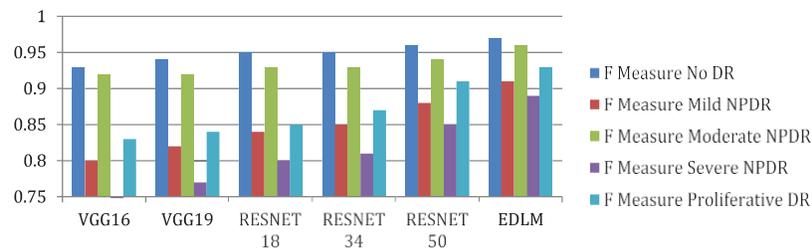

**Figure 10:** Comparison of F measure for enhanced deep learning model with other CNN models

**Table 6:** Computational time (in s) comparison

| Performance metrics | VGG16 | VGG19 | RESNET 18 | RESNET 34 | RESNET 50 | EDLM |
|---|---|---|---|---|---|---|
| Sensitivity | 23.24 | 23.07 | 23.43 | 24.23 | 19.87 | 17.01 |
| Specificity | 17.89 | 18.02 | 18.26 | 17.54 | 16.89 | 16.37 |
| F measure | 21.02 | 20.65 | 22.54 | 19.90 | 20.85 | 20.89 |

*Discussion*

The Rapid Diabetic Retinopathy (DR) Exposure (RDRE) system based on EDL has shown promising results in detecting diabetic retinopathy from fundus images. However, like any other technology, it also has its limitations. Some of the limitations of the EDL are: The accuracy of the proposed system is highly dependent on the quality of the fundus images. Poor quality images with low resolution, blur, or artifacts may result in inaccurate or false results. The proposed system can only detect diabetic retinopathy but cannot provide a complete diagnosis. Other ophthalmic conditions, such as glaucoma, macular degeneration, and cataracts, can also affect the retina, and the RDRE system may not be able to distinguish them.

## 6 Conclusions

Although the majority of work in the DR field is based on either disease finding or manual feature mining, this work aims to employ an EDLM framework for the automatic analysis of DR into its distinct classes. The dataset for the suggested model was generated from the publicly available machine learning repository, which in its pure form includes repetitive and superfluous characteristics. The dataset's required features were selected and extracted, but the Stochastic Gradient Descent (SGD) optimizer approach performed better. Initially, there is the application of deep CNN models for feature extraction. Afterward, it categorized the fundus images using a combination of the adopted models and the fully-connected layer. This feature combination scenario will involve diverse feature types, such as compactness, circularity, and roundness that are identified by the single shape descriptor. For dimensionality reduction, EDLM is used. The proposed EDLM architecture and the most recent deep CNN architectures like the ResNet and the VGGNet are used to classify and extract features from the exudate DR and the non-exudate DR. Results show that the EDL model has higher average sensitivity by 8.28% for VGG16, by 7.03% for VGG19, by 5.58% for ResNet18, by 4.26% for ResNet 34 and by 2.04% for ResNet 50 respectively. As a result, our suggested paradigm motivates academics to do comparable studies using high-dimensional data in a range of other health disciplines.

**Funding Statement:** The authors received no specific funding for this study.

**Conflicts of Interest:** The authors declare that they have no conflicts of interest to report regarding the present study.